\documentclass[runningheads]{llncs}
\usepackage{graphicx}
\usepackage{dingbat}
\usepackage{xcolor}
\usepackage{soul}
\usepackage{array}  
\usepackage{textcomp}
\usepackage{amsmath}
\usepackage{tikz}
\usetikzlibrary{arrows.meta, chains, fit, positioning}

\usepackage{makecell}
\usepackage[hidelinks]{hyperref}

\newcommand{\isep}{\mathrel{{.}\,{.}}\nobreak}
\newcommand{\smallfont}{\fontsize{1pt}{1.2pt}\selectfont}

\begin{document}
\title{DocParser: End-to-end OCR-free Information Extraction from Visually Rich Documents
}
\titlerunning{DocParser}
%
\author{Mohamed Dhouib\thanks{The corresponding author}\inst{1,2}\orcidID{0000-0002-5587-1028} \and
Ghassen Bettaieb\inst{1}\orcidID{0000-0003-3314-867X} \and
Aymen Shabou\inst{1}\orcidID{0000-0001-8933-7053}}

\authorrunning{M. Dhouib et al.}

\institute{DataLab Groupe, Credit Agricole S.A, Montrouge , France \and Ecole polytechnique, Palaiseau, France \\
\email{mohamed.dhouib@polytechnique.edu}$^{\ast}$
\email{\{ghassen.bettaieb,aymen.shabou\}@credit-agricole-sa.fr}  \\
\url{https://datalab-groupe.github.io/}\\
}

\maketitle              
\begin{abstract}
Information Extraction from visually rich documents is a challenging task that has gained a lot of attention in recent years due to its importance in several document-control based applications and its widespread commercial value. The majority of the research work conducted on this topic to date follow a two-step pipeline. First, they read the text using an off-the-shelf Optical Character Recognition (OCR) engine, then, they extract the fields of interest from the obtained text. The main drawback of these approaches is their dependence on an external OCR system, which can negatively impact both performance and computational speed. Recent OCR-free methods were proposed to address the previous issues. Inspired by their promising results, we propose in this paper an OCR-free end-to-end information extraction model named DocParser. It differs from prior end-to-end approaches by its ability to better extract discriminative character features. DocParser achieves state-of-the-art results on various datasets, while still being faster than previous works.

\end{abstract}
\section{Introduction}
Information extraction from visually rich documents (VRDs) is an important research topic that continues to be an active area of research \cite{chargrid,visualwordgrid,Cutie,cloudscan,layoutlm,docreader,trie++,Layout-aware,Graph_based-1} due to its importance in various real-world applications.

The majority of the existing information extraction from visually rich documents approaches \cite{layoutlm,Lambert,TILIT,Bros} depend on an external deep-learning-based Optical Character Recognition (OCR) \cite{text_detection,text_recognition} engine. They follow a two-step pipeline: First they read the text using an off-the-shelf OCR system then they extract the fields of interest from the OCR'ed text. These two-step approaches have significant limitations due to their dependence on an external OCR engine.
First of all, these approaches need positional annotations along with textual annotations for training.
Also, training an OCR model requires large scale datasets and huge computational resources. Using an external pre-trained OCR model is an option, which can degrade the whole model performance in the case of a domain shift. One way to tackle this is to fine-tune these off-the-shelf OCR models which is still a delicate task. In fact, the documents full annotations are generally needed to correctly fine-tune off-the-shelf OCR models, which is time-consuming and difficult to obtain.
OCR post-correction \cite{OCR_Post_Correction,OCR_Post_Correction_2} is an option to correct some of the recognition errors. However, this brings extra computational and maintenance cost.
Moreover, these two-step approaches rarely fully exploit the visual information because incorporating the textual information is already computationally expensive.

Recent end-to-end OCR-free information extraction approaches \cite{eaten,trie++,donut} were proposed  to tackle some of the limitations of OCR-dependant approaches. The majority of these approaches follow an encoder-decoder scheme. However, the used encoders are either unable to effectively model global dependence when they are primarily composed of  Convolutional neural network (CNN) blocks \cite{docreader,eaten} or they don't give enough privilege to  character-level features extraction when they are are primarily composed of Swin Transformer \cite{Swin} blocks \cite{donut,Dessurt}. In this paper, we argue that capturing both intra-character local patterns and inter-character long-range connections is essential for the information extraction task. The former is essential for character recognition and the latter plays a role in both the recognition and the localization of the fields of interest.

Motivated by the issues mentioned above, we propose an end-to-end OCR-free information extraction model named DocParser. DocParser has been designed in a way that allows it to efficiently perceive both intra-character patterns and inter-character dependencies.
Consequently, DocParser is up to two times faster than state-of-the-art methods while still achieving state-of-the-art results on various datasets.

\section{Related Work}
\subsection{OCR-dependant Approaches}
Most of the OCR-dependant approaches simply use an off-the-shelf OCR engine and only focus on the information extraction task. 

Prior to the development of deep learning techniques, earlier approaches \cite{earlier_approaches_0,earlier_approaches_1,earlier_approaches_2} either followed a probabilistic approach, relied on rules or used manually designed features which often results in failure when applied to unfamiliar templates. The initial deep learning approaches only relied on textual information and simply used pre-trained language models \cite{Bert,RoBERTa}. Later, several approaches tried to take the layout information into consideration. First, \cite{chargrid} proposed Chargrid, a new type of text representation that preserves the 2D layout of a document by encoding each document page as a two-dimensional grid of characters. Then, \cite{Bert_grid} added context to this representation by using a BERT language model. Later, \cite{visualwordgrid} improved the Chargrid model by also exploiting the visual information.
Graph-based models were also proposed to exploit both textual and visual information \cite{Graph_based-1,Graph_based-2}. 

To successfully model the interaction between the visual, textual and positional information, recent approaches \cite{layoutlm,Lambert,TILIT,Bros} resorted to pre-training large models. First \cite{layoutlmv0} tried to bring the success of large pre-trained language models into the multi-modal domain of document understanding and proposed LayoutLM.
LayoutLMv2 \cite{layoutlmv1} was later released where new pre-training tasks were introduced to better capture the cross-modality interaction in the pre-training stage. The architecture was also improved by introducing spatially biased attention and thus making the spatial information more influential. Inspired by the Vision Transformer (ViT) \cite{VIT}, \cite{layoutlm} modified LayoutLMv2 by using patch embeddings instead of a ResNeXt \cite{Resnext} Feature Pyramid Network \cite{FPN} visual backbone and released LayoutLMv3. Pre-training tasks were also improved compared to previous versions. \cite{Lambert} proposed LAMBERT which used a modified RoBERTa \cite{RoBERTa} that also exploits the layout features obtained from an OCR system. \cite{TILIT} proposed TILT, a pre-trained encoder-decoder model. \cite{Bros} tried to fully exploit the textual and layout information and released Bros which achieves good results without relying on the visual features. However, the efficiency and the computational cost of all the previously cited works are still hugely impacted by the used OCR system. 

\subsection{End-to-end Approaches}
In recent years, end-to-end approaches were proposed for the information extraction task among many other Visually-Rich Document Understanding (VRDU) tasks.
 \cite{eaten,docreader} both used a CNN-based encoder and a recurrent neuronal network coupled with an attention mechanism decoder. However, the accuracy of these two approaches is limited and they perform relatively badly on small datasets.
\cite{trie++} proposed TRIE++, a model that learns simultaneously both the text reading and the information extraction tasks via a multi-modal context block that bridges the visual and natural language processing tasks. \cite{VIES} released VIES which simultaneously performs text detection, recognition and information extraction.
However, both TRIE++ and VIES require the full document annotation to be trained. \cite{donut} proposed Donut, an encoder-decoder architecture that consists of a Swin Transformer \cite{Swin} encoder and a Bart \cite{Bart}-like decoder. \cite{Dessurt} released Dessurt, a model that processes three streams of tokens, representing visual tokens, query tokens and the response. Cross-attention is applied across different streams to allow them to share and transfer information into the response. To process the visual tokens, Dessurt uses a modified Swin windowed attention that is allowed to attend to the query tokens. Donut and Dessurt achieved promising results, however, they don't give enough privilege to local character patterns which leads to sub-optimal results for the information extraction task.
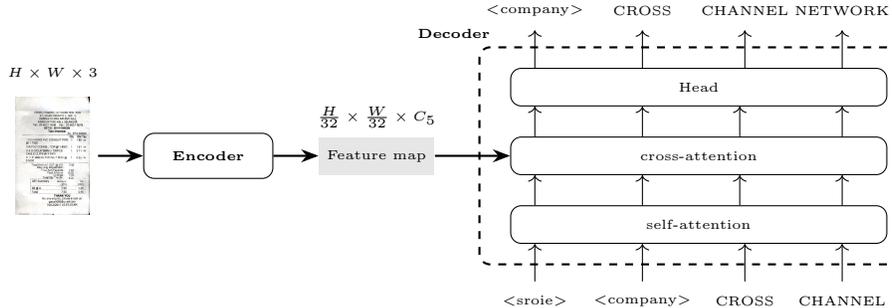
\begin{figure}
\tikzset{
    *|/.style={
        to path={
            (perpendicular cs: horizontal line through={(\tikztostart)},
                                 vertical line through={(\tikztotarget)})
            -- (\tikztotarget) \tikztonodes
        }
    }
}
\begin{tikzpicture}[
image/.style = {rectangle, minimum height=.5cm, minimum width=.5cm, align=flush center},
encoder/.style = {rounded corners, draw, 
                  text width=1.5cm, align=center,
                  minimum height=.6cm,
                  minimum width=1.5cm},
decoder/.style = {rounded corners, draw, 
                  minimum height=.5cm,
                  minimum width=5cm,
                  text width=3cm, align=flush center, 
                  line join=round
                  },
inouts/.style = {align=flush center,
                  minimum height=.5cm,
                  minimum width=.5cm},
connection/.style = {thick,-Stealth},
                    ]
\node(n1) [image, label=above:{\tiny $H \times W \times 3$}] {\includegraphics[width=.08\textwidth]{figures/X51005230657.jpg}};
\node(n2) [encoder, right=.6cm of n1] {\tiny \textbf{Encoder}};
\node(n3) [image, right=.6cm of n2.east, label=above:{\tiny $\frac{H}{32}\times \frac{W}{32}\times C_5$}, fill=gray!20] {\tiny Feature map};
\node(n4) [decoder, right=1cm of n3.east]{\tiny cross-attention};
\node(n5) [decoder, above=.4cm of n4.north]{\tiny Head};
\node(n6) [decoder, below=.4cm of n4.south]{\tiny self-attention};

\draw[->,connection] (n1.east) to (n2.west);
\draw[->,connection] (n2.east) to (n3.west);
\draw[->,connection] (n3.east) to (n4.west);

\node(in1) [inouts, xshift=4.5ex, below=.5cm of n6] {\tiny CROSS};
\node(in2) [inouts,  xshift=-5.5ex, below=.5cm of n6]  {\tiny $<$company$>$};
\node(in3) [inouts,  xshift=-16ex, below=.5cm of n6] {\tiny $<$sroie$>$};
\node(in4) [inouts, xshift=14ex, below=.5cm of n6] {\tiny CHANNEL};

\draw[<-,*|]  (n6.south) to[*|] (in3.north);
\draw[<-]   (n6.south) to[*|] (in2.north);
\draw[<-]   (n6.south) to[*|] (in1.north);
\draw[<-]   (n6.south) to[*|] (in4.north);

\draw[->] ([xshift=4.5ex]n6.north) to[*|] ([xshift= 4ex]n4.south);
\draw[->] ([xshift=-5.5ex]n6.north) to[*|] ([xshift= -5.5ex]n4.south);
\draw[->] ([xshift=-16ex]n6.north) to[*|] ([xshift= -16ex]n4.south);
\draw[->] ([xshift=12ex]n6.north) to[*|] ([xshift= 14ex]n4.south);

\draw[->] ([xshift=4.5ex]n4.north) to[*|] ([xshift= 4ex]n5.south);
\draw[->] ([xshift=-5.5ex]n4.north) to[*|] ([xshift= -5.5ex]n5.south);
\draw[->] ([xshift=-16ex]n4.north) to[*|] ([xshift= -16ex]n5.south);
\draw[->] ([xshift=12ex]n4.north) to[*|] ([xshift= 14ex]n5.south);

\node(out1)  [inouts,xshift=-5.5ex, above=.5cm of n5] {\tiny CROSS};
\node(out2) [inouts,xshift=-16ex,above=.5cm of n5]  {\tiny $<$company$>$};
\node(out4) [inouts,xshift=4.5ex,above=.5cm of n5] {\tiny CHANNEL};
\node(out3) [inouts, xshift=14ex,above=.5cm of n5] {\tiny NETWORK};

\draw[->,*|]  (n5.north) to (out2.south);
\draw[->]   (n5.north) to[*|] (out1.south);
\draw[->]   (n5.north) to[*|] (out3.south);
\draw[->]   (n5.north) to[*|] (out4.south);

\node[draw, rounded corners, thick, dashed, inner ysep=2ex, inner xsep=2ex, xshift=-1ex, fit=(n4) (n5) (n6), label=150:{\tiny \textbf{Decoder}}] (box) {};

\end{tikzpicture}
\caption{ \textbf{An overview of DocParser's architecture}. The input image of size $H \times W \times 3$ is first encoded to generate a feature map of size $\frac{H}{32}\times \frac{W}{32}\times C_5$ containing relevant visual information. The feature map is then fed to the decoder along with a task token to auto-regressively generate tokens that represent the fields of interest. For the purpose of simplification, the figure does not include residual connections and the feed-forward sub-layer.}
\label{fig:docparser_overview}
\end{figure}
\section{Proposed Method}
This section introduces DocParser, our proposed end-to-end information extraction from VRDs model.

Given a document image and a task token that determines the fields of interest, DocParser produces a series of tokens representing the extracted fields from the input image. DocParser architecture consists of a visual encoder followed by a textual decoder.
An overview of DocParser’s architecture is shown on figure \ref{fig:docparser_overview}.
The encoder consists of a three-stage progressively decreased height convolutional neural network that aims to extract intra-character local patterns, followed by a three-stage progressively decreased width Swin Transformer \cite{Swin} that aims to capture long-range dependencies.
The decoder consists of $n$ Transformer layers. Each layer is principally composed of a multi-head self-attention sub-layer followed by a multi-head cross-attention sub-layer and a feed-forward sub-layer as explained in \cite{attention}.

\subsection{Encoder} 
\begin{figure}
\centering
\begin{tikzpicture}[
start chain = going right,
windowedsmall/.style = {rounded corners, draw, 
                  text width=0.6cm,align=flush center, 
                  on chain, join=by line,
                  minimum height=3.25cm,
                  minimum width=0.7cm,
                  inner ysep=7.5ex},
windowedbig/.style = {rounded corners, draw, 
                  text width=0.7cm,
                  align=flush center, 
                  on chain, join=by line,
                  minimum height=3.25cm,
                  minimum width=0.7cm,
                  inner ysep=7.5ex},
convnext/.style = {rounded corners, draw, 
                  text width=0.5cm,align=flush center, 
                  on chain, join=by line,
                  minimum height=3.25cm,
                  minimum width=0.6cm,
                  inner ysep=7.5ex},
vertical/.style = {rounded corners, draw, 
                  text width=3cm, align=flush center, 
                  minimum height=.05cm,
                  minimum width=3cm,
                  on chain, join=by line},
stage/.style={draw, rounded corners, thick, dashed, inner ysep=4ex, inner xsep=2ex, yshift=1ex},
line/.style = {-Stealth}
                    ]
\node (n1) [draw, rotate=90,anchor=north, rounded corners, on chain, join=by line, minimum height=.4cm, minimum width=1cm, label=0:{ \smallfont  $H$.$W$.$3$}] {\smallfont Image};
\node (n2) [vertical, rotate=90, right=.25cm of n1.south, anchor=north] {\smallfont Patch Embedding};

\node (n3) [convnext, right=.3cm of n2.south, label=270:{$\times3$}, label=88:{\scriptsize Stage 1}] {\smallfont CN \\  $(7,7)$};
\node (n4) [vertical, rotate=90, right=.2cm of n3, anchor=north] {\smallfont downsampling Layer}; 

\node (n5) [convnext, right=.4cm of n4.south, label=270:{$\times6$}, label=88:{\scriptsize Stage 2}] {\smallfont CN\\  $(7,7)$};
\node (n6) [vertical, rotate=90, right=.2cm of n5, anchor=north] {\smallfont downsampling Layer};

\node (n7) [convnext, right=.4cm of n6.south, label=270:{$\times6$}, label=88:{\scriptsize Stage 3}] {\smallfont CN \\  $(7,7)$};
\node (n8) [vertical, rotate=90, right=.2cm of n7, anchor=north] {\smallfont downsampling Layer};

\node (n9) [windowedsmall,  right=.4cm of n8.south, label=270:{$\times2$}, label=88:{\scriptsize Stage 4}] {\smallfont Swin \\   $(5,40)$};
\node (n10) [vertical, rotate=90, right=.2cm of n9, anchor=north] {\smallfont downsampling Layer};

\node (n11) [windowedsmall, right=.4cm of n10.south, label=270:{$\times2$}, label=88:{\scriptsize Stage 5}] {\smallfont Swin \\  $(5,20)$};
\node (n12) [vertical, rotate=90, right=.2cm of n11, anchor=north] {\smallfont downsampling Layer};

\node (n13) [windowedbig, right=.4cm of n12.south, label=270:{$\times2$},label=88:{\scriptsize Stage 6}] {\smallfont Swin \\$(10,10)$};

\node [draw, rounded corners, thick, dashed, align=center, inner ysep=5ex, inner xsep=1ex, yshift=0ex, fit=(n3) (n4), label=100:{\smallfont $\frac{H}{4}.\frac{W}{4}.C_0$}] (box) {};

\node[draw, rounded corners, thick, dashed, inner ysep=5ex, inner xsep=1ex, fit=(n5) (n6), label=100:{\smallfont $\frac{H}{8}.\frac{W}{4}.C_1$}] (box) {};

\node[draw, rounded corners, thick, dashed, inner ysep=5ex, inner xsep=1ex, fit=(n7) (n8), label=100:{\smallfont $\frac{H}{16}.\frac{W}{4}.C_2$}] (box) {};

\node[draw, rounded corners, thick, dashed, inner ysep=5ex, inner xsep=1ex, fit=(n9) (n10), label=100:{\smallfont $\frac{H}{32}.\frac{W}{8}.C_3$}] (box) {};

\node[draw, rounded corners, thick, dashed, inner ysep=5ex, inner xsep=1ex, fit=(n11) (n12), label=100:{\smallfont $\frac{H}{32}.\frac{W}{16}.C_4$}, label=87:{\smallfont $\frac{H}{32}.\frac{W}{32}.C_5$}] (box) {};
\node[draw, rounded corners, thick, dashed, inner ysep=5ex, inner xsep=1ex, fit=(n13), label=87:{\smallfont $\frac{H}{32} .\frac{W}{32}.C_5$}] (box) {};

\end{tikzpicture}
\caption{ \textbf{The architecture of DocParser's encoder.} $H$ and $W$ represent the height and the width of the input image. $C_i,$ $ i \in [0..5]$ represent the number of channels at different stages. The first three stages are composed of \textbf{C}onv\textbf{N}ext (CN) blocks with a filter size of $(7, 7)$. The last three stages are composed of Swin Transformer blocks with different attention window sizes. The windows' heights and widths are respectively equal to (5,40), (5,20) and (10,10).}
\label{fig:encoder_architecture}
\end{figure}
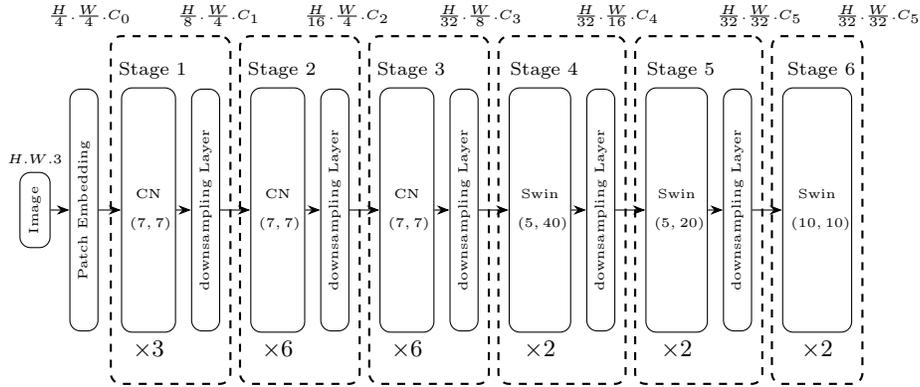

The encoder is composed of six stages.
The input of the encoder is an image of size $H \times W \times 3$. It is first transformed to $\frac{H}{4} \times \frac{W}{4}$
patches of dimension $C_0$ via an initial patch embedding. Each patch either represents a fraction of a text character or a fraction of a non-text component of the input image. First, three stages composed of ConvNext \cite{ConvNext} blocks are applied at different scales for character-level discriminative features extraction. Then three stages of Swin Transformer blocks are applied with varying window sizes in order to capture long-range dependencies. The output of the encoder is a feature map of size $\frac{H}{32} \times \frac{W}{32} \times C_5$ that contains multi-grained features. An overview of the encoder architecture is illustrated in figure \ref{fig:encoder_architecture}.

\subsubsection{Patch Embedding}\hfill\\
\\
Similar to \cite{SVTR}, we use a progressive overlapping patch embedding. For an input image of size $W \times H \times 3$, a $3 \times 3$ convolution with stride $2$ is first applied to have an output of size $\frac{W}{2} \times \frac{H}{2} \times \frac{C_0}{2}$. It is then followed by a  normalization layer and another $3 \times 3$ convolution with stride $2$. The size of the final output is $\frac{W}{4} \times \frac{H}{4} \times C_0$.

\subsubsection{ConvNext-based Stages}\hfill \\
\\
The first three stages of DocParser's encoder are composed of ConvNext blocks. Each stage is composed of several blocks. The kernel size is set to $7$ for all stages. At the end of each stage, the height of the feature map is reduced by a factor of two and the number of channels $C_i,$ $ i \in [1,2,3]$ is increased to compensate for the information loss. The feature map width is also reduced by a factor of two at the end of the third stage. The role of these blocks is to capture the correlation between the different parts of each single character and to encode the non-textual parts of the image. 
We don't reduce the width of the feature map between these blocks in order to avoid encoding components of different characters in the same feature vector and thus allowing discriminative character features computation. We note that contrary to the first encoder stages where low-level features extraction occurs, encoding components of different characters in the same feature vector doesn't affect performance if done in the encoder last stages where high-level features are constructed. This is empirically demonstrated in section \ref{abla}.
We chose to use convolutional blocks for the early stages mainly due to their good ability at modeling local correlation at a low computational cost.

\subsubsection{Swin Transformer-based Stages}\hfill\\
\\
The last three stages of the encoder are composed of Swin Transformer blocks. We modify Swin's window-based multi-head self-attention to be able to use rectangular attention windows. At the output of the fourth and fifth stages, the width of the feature map is reduced by a factor of two and the number of channels is increased to compensate for the information loss.
The role of these layers is to capture the correlation between the different characters of the input image or between textual and non-textual components of the image.
In the forth and fifth stage, the encoder focuses on capturing the correlation between characters that belong to adjacent sentences. This is accomplished through the use of horizontally wide windows, as text in documents typically has an horizontal orientation.
In the last stage, the encoder focuses on capturing long-range context in both directions. This is achieved through the use of square attention windows.
As a result, the output of the encoder is composed of multi-grained features that not only encode intra-character local patterns which are essential to distinguish characters but also capture the correlation between textual and non-textual components which is necessary to correctly locate the fields of interest. We note that positional embedding is added to the encoder's feature map before the encoder's forth stage.

\subsection{Decoder} 
\setcounter{footnote}{0} \noindent
The decoder takes as input the encoder's output and a task token. It then outputs autoregressively  several tokens that represent the fields of interest specified by the input token. The decoder consists of $n$\footnote{For our final model, we set $n$=1} layers, each one is similar to a vanilla Transformer decoder layer. It consists of a multi-head self-attention sub-layer followed by a multi-head cross-attention sub-layer and a feed-forward sub-layer.

\subsubsection{Tokenization}
\begin{figure}
\tikzset{
    *|/.style={
        to path={
            (perpendicular cs: horizontal line through={(\tikztostart)},
                                 vertical line through={(\tikztotarget)})
            -- (\tikztotarget) \tikztonodes
        }
    }
}
\begin{center}
\begin{tikzpicture}
\node (n1) [draw, rounded corners, text width=11cm] { \small
\{ \\
\textcolor{red}{"menu"}: [
\\  \{\textcolor{blue}{"name"}: "Macchiato", \textcolor{violet}{"price"}: "17500"\}, 
\\ \{\textcolor{blue}{"name"}: "TEA", \textcolor{violet}{"price"}: "5000"\}, 
\\],
\\ \textcolor{gray} {"total"}: \{\textcolor{brown}{"total-price"}: "22500", \textcolor{olive}{"cash-price"}: "22500"\} 
\\ \} 
}; 
\newcommand{\autour}[1]{\tikz[baseline=(X.base)]\node [draw=blue,fill=blue!5,semithick,rectangle,inner sep=2pt, rounded corners=3pt] (X) {#1};}

\node (n2) [text width=11cm, draw, rounded corners, below=1cm of n1.south] { \small
\textcolor{purple}{ \autour{\textless Task \textgreater }}
\textcolor{red}{ \autour{\textless menu \textgreater }}
\textcolor{orange}{\autour{\textless item \textgreater} }
\textcolor{blue}{\autour{\textless name \textgreater}}  \autour{Mac} \autour{chi} \autour{ato} \textcolor{blue}{\autour{\textless name /\textgreater}}
\textcolor{violet}{\autour{\textless price \textgreater}}  \autour{17} \autour{500}  \textcolor{violet}{\autour{\textless price /\textgreater}}
\textcolor{orange}{\autour{\textless item /\textgreater}}
\textcolor{orange}{\autour{\textless item \textgreater }}
\textcolor{blue}{\autour{\textless name \textgreater}} \autour{TEA}  \textcolor{blue}{\autour{\textless name /\textgreater}}
\textcolor{violet}{\autour{\textless price \textgreater}}  \autour{5000}  \textcolor{violet}{\autour{\textless price /\textgreater}}
\textcolor{orange}{\autour{\textless item /\textgreater}}
\textcolor{red} {\autour{\textless menu /\textgreater}}
\textcolor{gray} {\autour{\textless total \textgreater} }
\textcolor{brown}{\autour{\textless total-price \textgreater}}  \autour{22} \autour{500} \textcolor{brown}{\autour{\textless total-price /\textgreater }}
\textcolor{olive}{\autour{\textless cash-price \textgreater}}  \autour{22} \autour{500} \textcolor{olive}{\autour{\textless cash-price /\textgreater}}
\textcolor{gray} {\autour{\textless total /\textgreater}} \textcolor{purple}{ \autour{\textless End \textgreater }}
};

\draw[->,-Stealth] (n1.south) to (n2.north);

\end{tikzpicture}
\end{center}
\caption{ \textbf{An illustration of the processing applied to the textual ground-truth.} }
\label{fig:token}
\end{figure}
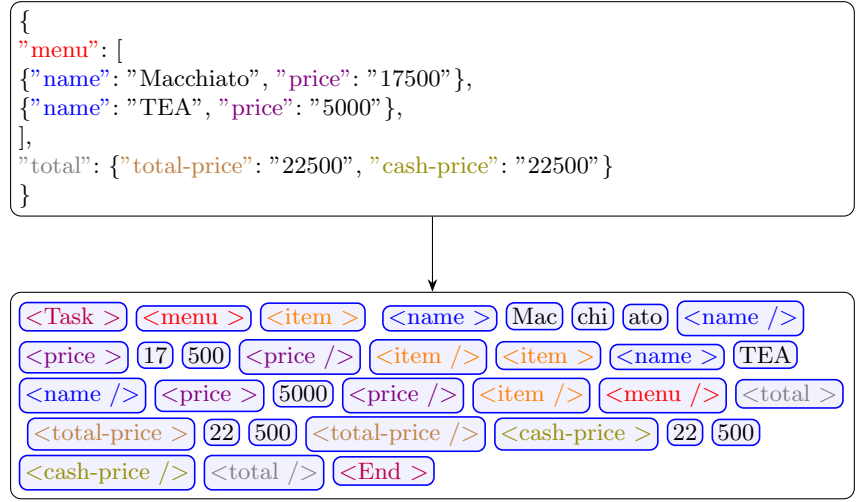

We use the tokenizer of the RoBERTa model \cite{RoBERTa} to transform the ground-truth text into tokens. This allows to reduce the number of generated tokens, and so the memory consumption as well as training and inference times, while not affecting the model performance as shown in section \ref{abla}.
Similar to \cite{donut}, special tokens are added to mark the start and the end of each field or group of fields. Two additional special tokens $<item>$ and $<item/>$  are used to separate fields or group of fields appearing more than once in the ground truth. An example is shown in figure \ref{fig:token}.

\subsubsection{At Training Time}
When training the model, we use a teacher forcing strategy. This means that we give the decoder all the ground truth tokens as input. Each input token corresponding last hidden state is used to predict the next token. To ensure that each token only attends to previous tokens in the self-attention layer, we use a triangular attention mask that masks the following tokens.

\section{Expriments and Results}
\subsection{Pre-training}

We pre-train our model on two different steps :
\subsubsection{Knowledge Transfer Step}Using an $L2$ Loss, we teach the ConvNext-based encoder blocks to produce the same feature map as the PP-OCR-V2 \cite{Paddle} recognition backbone which is an enhanced version of MobileNetV1 \cite{mobilenet}. A pointwise convolution is applied to the output of the ConvNext-based blocks in order to obtain the same number of channels as the output of PP-OCR-V2 recognition backbone. The goal of this step is to give the encoder the ability to extract discriminative intra-character features. We use 0.2 million documents from the IIT-CDIP \cite{CDIP} dataset for this task. We note that even though PP-OCR-V2 recognition network was trained on text crops, the features generated by its backbone on a full image are still useful thanks to the translation equivariance of CNNs.
\subsubsection{Masked Document Reading Step} After the knowledge transfer step, we pre-train our model on the task of document reading. In this pre-training phase, the model learns to predict the next textual token while conditioning on the previous textual tokens and the input image. To encourage joint reasoning, we mask several $32 \times 32$ blocks representing approximately fifteen percent of the input image. In fact, in order to predict the text situated within the masked regions, the model is obliged to understand its textual context. As a result, DocParser learns simultaneously to recognize characters and the underlying language knowledge.
We use 1.5 million IIT-CDIP documents for this task. These documents were annotated using Donut. Regex rules were applied to identify poorly read documents, which were discarded.

\subsection{Fine-tuning}
\begin{figure}
    \includegraphics[width=.33\textwidth]{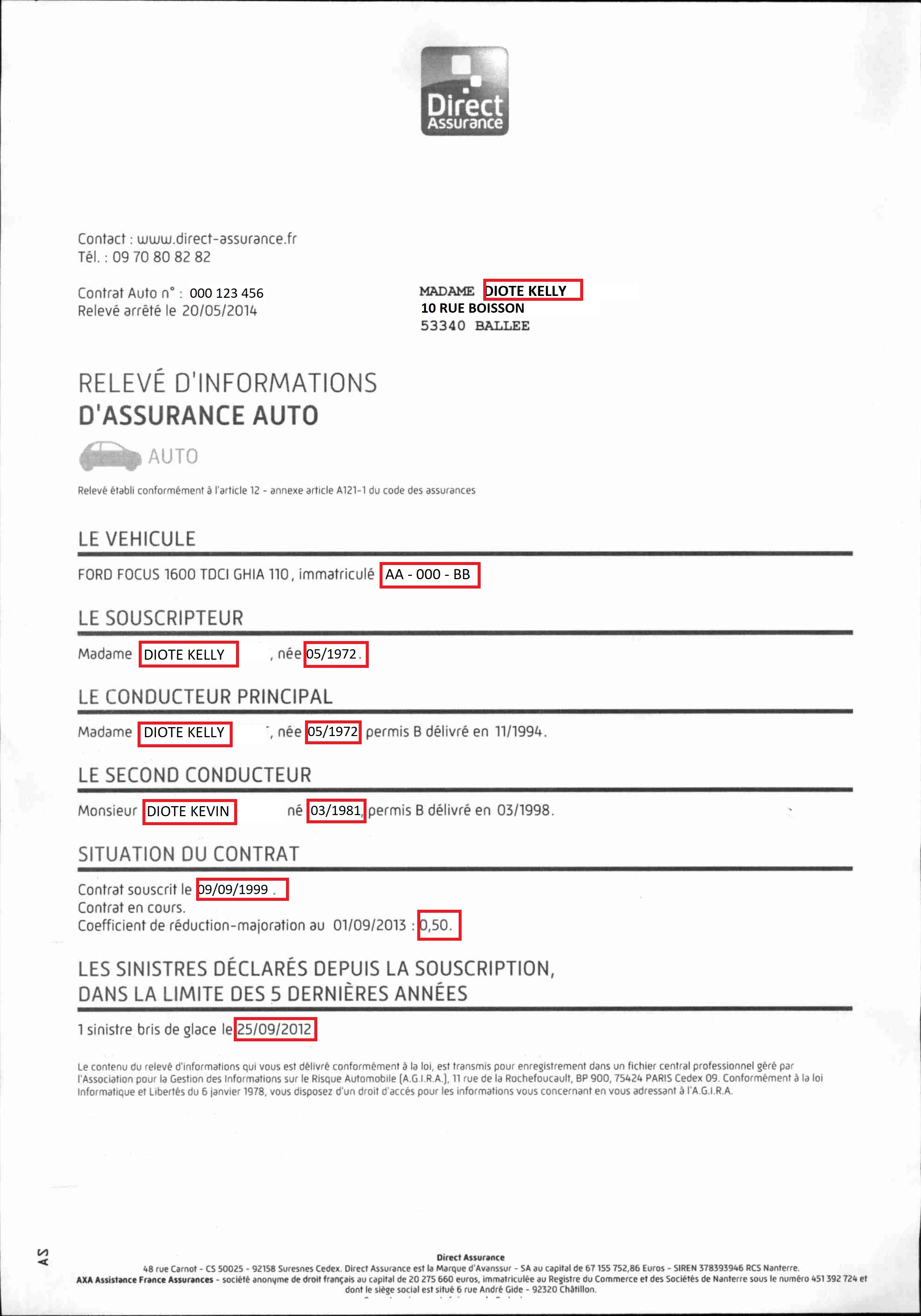}\hfill
    \includegraphics[width=.33\textwidth]{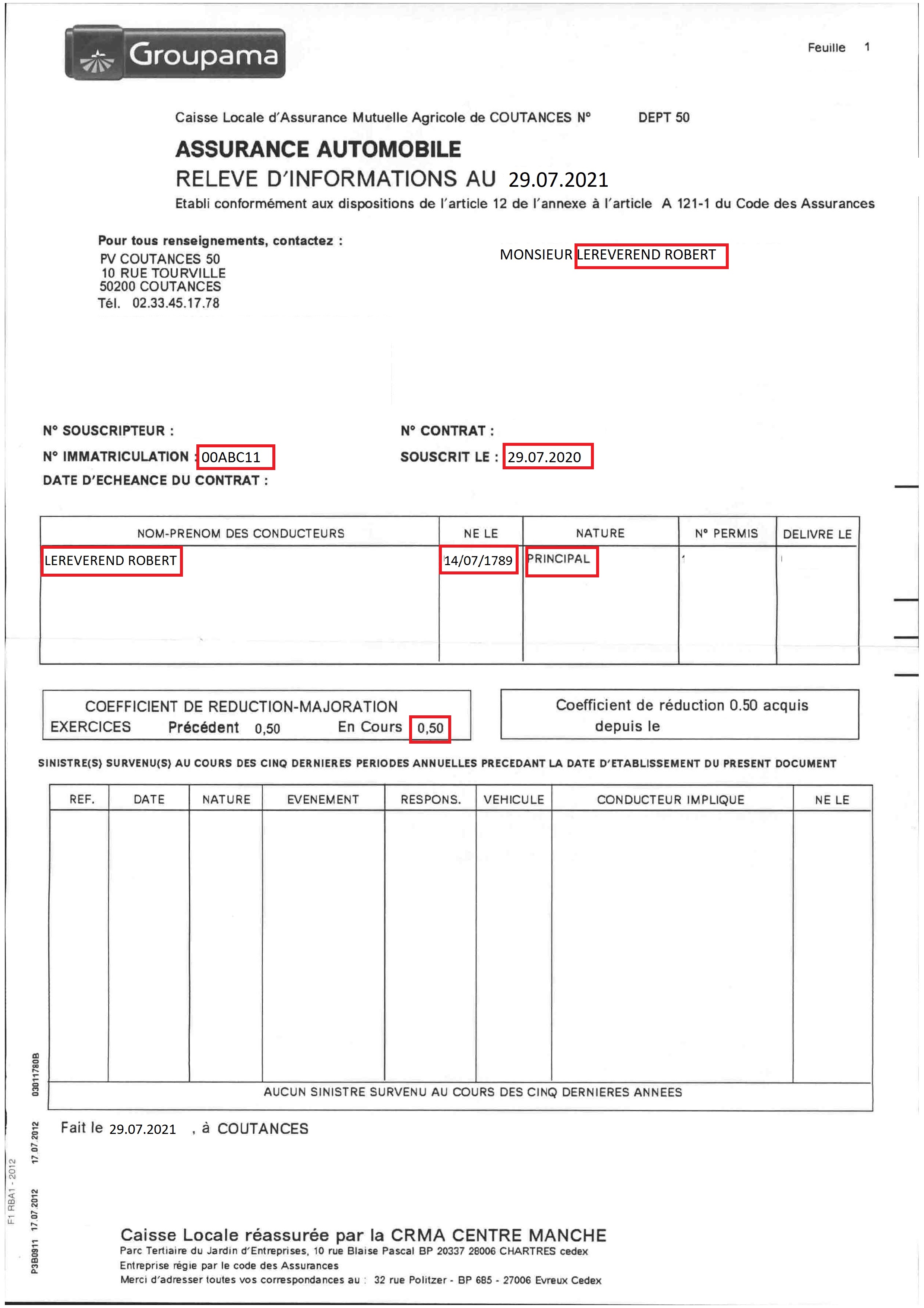} \hfill
    \includegraphics[width=.33\textwidth]{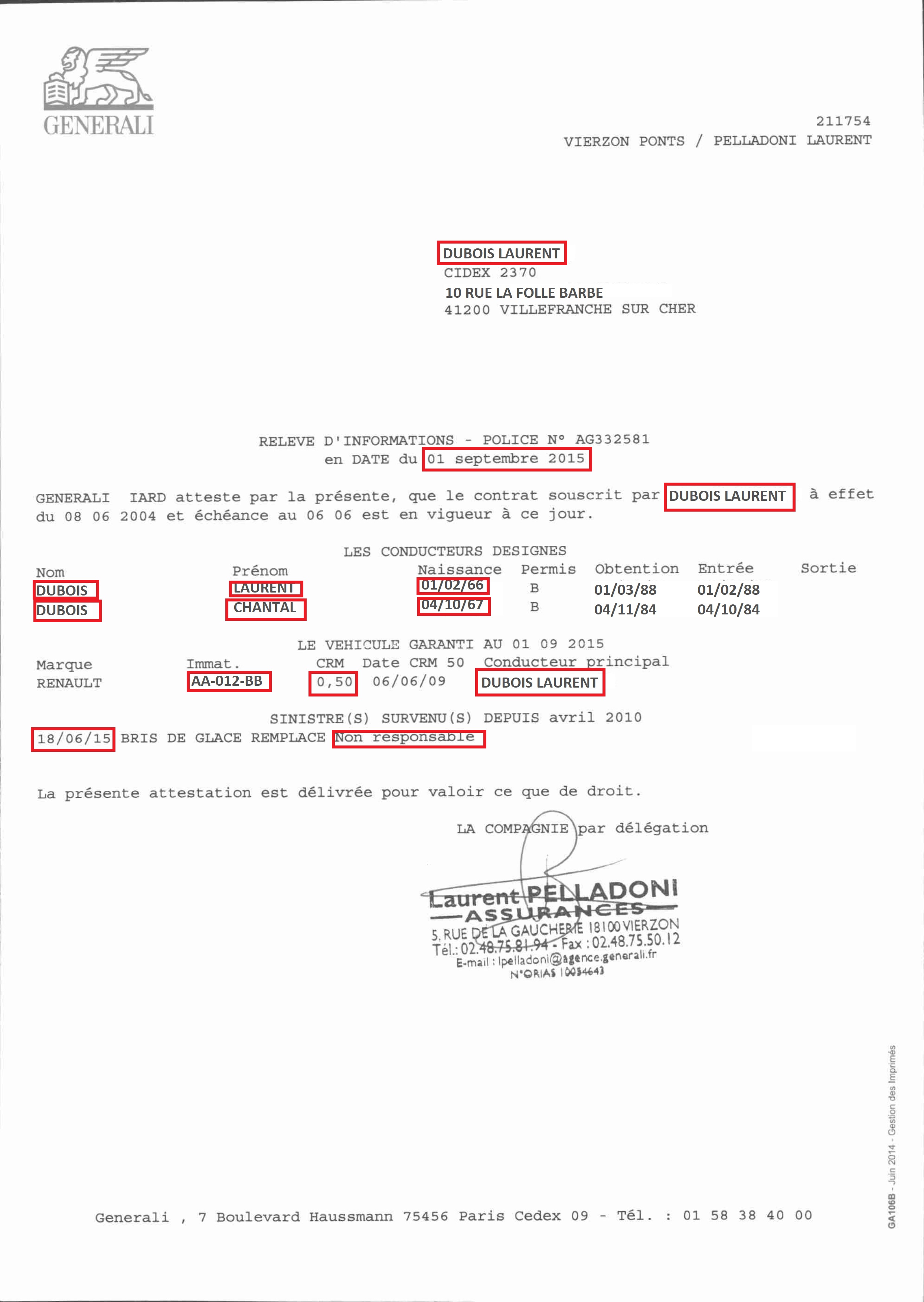}\hfill

    \caption{ \textbf{Anonymized samples from our private in-house dataset.} The fields of interest are located within the red boxes.}\label{fig:in-house_dataset}
    \label{fig:inhouse_dataset}
\end{figure}

After the pre-training stage, the model is fine-tuned on the information extraction task. We fine-tune DocParser on three datasets: SROIE and CORD which are public datasets and an in-house private Information Statement Dataset.
\paragraph{SROIE}: A public receipts dataset with 4 annotated unique fields : company, date, address, and total. It contains 626 receipts for
training and 347 receipts for testing.
\paragraph{CORD}: A public receipts dataset with 30 annotated unique fields of interest. It consists of 800 train, 100 validation and 100 test receipt images.
\paragraph{Information Statement Dataset (ISD) }: A private information statement dataset with 18 annotated unique fields of interest. It consists of 7500 train, 3250 test and 3250 eval images. The documents come from 15 different insurers, each insurer has around 4 different templates. We note that for the same template, the structure can vary depending on the available information. On figure \ref{fig:inhouse_dataset} we show 3 samples from 3 different insurers.
\newline

\subsection{Evaluation Metrics}
We evaluate our model using two metrics:
\subsubsection{Field-level F1 Score}
The field-level F1 score checks whether each
extracted field corresponds exactly to its value in the ground truth. For a given field, the field-level F1 score assumes that the extraction has failed even if one single character is poorly predicted. The field-level F1 score is described using the field-level precision and recall as:
\begin{equation}
\text{Precision} = \frac{\text{The number of exact field matches}}{\text{The number of the detected fields}}
\end{equation}
\begin{equation}
\text{Recall} = \frac{ \text{The number of exact field matches}}{\text{The number of  the ground truth fields}}
\end{equation}
\begin{equation}
\text{F1} = \frac{ 2 \times \text{Precision} \times \text{Recall}}{\text{Precision} + \text{Recall}}
\end{equation}

\subsubsection{Document Accuracy Rate (DAR)}
This metric evaluates the number of documents that are completely and correctly processed by the model. If for a given document we have even one false positive or false negative, the DAR assumes that the extraction has failed. 
This metric is a challenging one, but requested in various industrial applications where we need to evaluate at which extent the process is fully automatable.

\subsection{Setups}
 The dimension of the input patches and the output vectors of every stage $C_i,$ $ i \in [0\isep 5]$ are respectively set to $64$, $128$, $256$, $512$, $768$, and $1024$. We set the number of decoder layers to $1$. This choice is explained in Section~\ref{abla}. For both pre-training and fine-tuning we use the Cross-Entropy Loss, AdamW \cite{ADAMW} optimizer with weight decay of $0.01$ and stochastic depth \cite{stocha} with a probability equal to $0.1$. We also follow a light data augmentation strategy which consists of light re-scaling and rotation as well as brightness, saturation, and contrast augmentation applied to the input image. For the pre-training phase, we set the input image size to $2560 \times 1920$. The learning rate is set to $1e-4$. 
The pre-training is done on 7 A100 GPUs with a batch size of 4 on each GPU. We use gradient accumulation of 10 iterations, leading
to an effective batch size of $280$. For the fine-tuning, the resolution is set to $1600 \times 960$ for CORD and SROIE datasets and $1600 \times 1280$ for the Information Statement Dataset. We pad the input image in order to maintain its aspect ratio. We also use a Cosine Annealing scheduler \cite{cosine} with an initial learning rate of $3e-5$ and a batch size of $8$.
 
\subsection{Results}
\begin{table}[htbp]
\setlength\tabcolsep{0pt}
\setlength\extrarowheight{2pt}
\caption{  \textbf{Performance comparisons on the three datasets.} The field-level F1-score and the extraction time per image on an Intel Xenion W-2235 CPU are reported. In order to ensure a fair comparison, we exclude parameters related to vocabulary. Additional parameters $\alpha^{*}$ and time $t^{*}$ for the OCR step should be considered for LayouLM-v3. For the ISD dataset $t^{*}$ is equal to 3.6 seconds.}
\label{tab2}
\begin{tabular*}{\textwidth}{@{\extracolsep{\fill}}*{11}{c}}
\hline
\multicolumn{3}{c}{ } & \multicolumn{2}{c}{SROIE} & \multicolumn{2}{c}{CORD} & \multicolumn{2}{c}{ISD} \\
\cline{4-5} \cline{6-7}\cline{8-9}\\
 &  OCR & Params(M) &F1(\%)& Time(s) & F1(\%)& Time(s)& F1(\%)& Time(s)  \\
\hline
LayoutLM-v3 &  \checkmark & $87+\alpha^{*}$ & $77.7$ & $2.1 + t^{*}$  & $80.2$ & $2.1+t^{*}$& $90.8$ & $4.1+t^{*}$\\
Donut &   & 149 & 81.7 & 5.3  & 84 &  5.7 & 95.4 & 6.7\\
Dessurt &   & 87 & 84.9 & 16.7  & 82.5 &  17.9 & 93.5 & 18.1\\
{\bfseries DocParser} &   & \textbf{70} & \textbf{87.3} & 3.5 & \textbf{84.5} & 3.7 & \textbf{96.2} & \textbf{4.4} \\
\hline
\end{tabular*}
\end{table}

We compare DocParser to Donut, Dessurt and LayoutLM-v3. The results are summarized in table \ref{tab2}. A comparison of inference speed on an NVIDIA Quadro RTX 6000 GPU is presented in table \ref{tabgpu}. Per-field extraction performances on our Information Statement Dataset can be found in table \ref{tab3}. DocParser achieves a new state-of-the-art on SROIE, CORD and our Information Statement Dataset with an improvement of respectively 2.4, 0.5 and 0.8 points over the previous state-of-the-art. In addition, Docparser has a significantly faster inference speed and less parameters.

\setlength{\tabcolsep}{6pt}
\begin{table}
\centering
\setlength\extrarowheight{2pt}
\caption{\textbf{Comparison of inference speed on GPU.} Extraction time (seconds) per image on an  NVIDIA Quadro RTX 6000 GPU is reported. Additional time $t^{*}$ for the OCR step should be considered for LayouLM-v3. For the ISD dataset $t^{*}$ is equal to 0.5 seconds.}\label{tempsgpu}
\label{tabgpu}
\begin{tabular}{l l l l l}
\hline
 &  SROIE & CORD& ISD \\
\hline
LayoutLM-v3 & 0.041 + $t^{*}$ & 0.039 + $t^{*}$ & 0.065 + $t^{*}$\\
Donut & 0.38 &0.44 & 0.5\\
Dessurt & 1.2 &1.37 & 1.39\\
{\bfseries DocParser} & 0.21 &0.24 & \textbf{0.25}\\
\end{tabular}
\end{table}

\setlength{\tabcolsep}{8pt}
\label{OURDATASET}
\begin{table}
\centering
\caption{\textbf{Extraction performances on our Information Statement Dataset.} Per field (field-level) F1-score, field-level F1-score mean, DAR, and extraction time per image on an Intel Xenion W-2235 CPU are reported. The OCR engine inference time $t^{*}$ should be considered for LayouLM-v3.}\label{tab3}
\begin{tabular}{l | l l l l }
\hline
\thead{Fields}  & LayoutLM & DONUT & Dessurt & \bfseries{DocParser}  \\
\hline
\thead{Name of first driver} & \thead{85.7} & \thead{91.3} & \thead{89.7} & \thead{\textbf{92.9}} \\
\thead{Name of second driver} & \thead{82.7} & \thead{90.1} & \thead{89.1} & \thead{\textbf{92.1}} \\
\thead{Name of third driver} & \thead{76.8} & \thead{94.2} & \thead{91.7} & \thead{\textbf{94.7}} \\
\thead{Bonus-Malus} & \thead{96} & \thead{98.5} & \thead{98.1} & \thead{\textbf{99}} \\
\thead{Date of birth of \\ first driver} & \thead{97.1} & \thead{97.1} & \thead{97.3} & \thead{\textbf{98.3}} \\
\thead{Date of birth of \\ second driver}& \thead{95.8} & \thead{96.9} &  \thead{96.3} & \thead{\textbf{97.2}}  \\
\thead{Date of birth of \\ third driver} & \thead{91.5} & \thead{\textbf{93.2}} & \thead{90.8} &\thead{92.8} \\
\thead{Date of termination \\ of contract} & \thead{92} & \thead{97} &  \thead{95.7} &\thead{\textbf{97.8}} \\
\thead{Date of the first accident} & \thead{95.6} & \thead{96} & \thead{96.3} & \thead{\textbf{97.2}} \\
\thead{Date of the second accident} & \thead{94.2} & \thead{95.2} & \thead{94.8} & \thead{\textbf{96.1}} \\
\thead{Subscription date} & \thead{94.7} & \thead{97.9} & \thead{95.7} & \thead{\textbf{98.5}} \\
\thead{Date of creation of \\ the document} & \thead{95.6} & \thead{97.5} &  \thead{96.8} &\thead{\textbf{98}} \\
\thead{Matriculation} & \thead{95.2} & \thead{94} & \thead{94.6} &\thead{\textbf{96.7}} \\
\thead{Name of first accident's \\ driver} & \thead{85.9} & \thead{85.9} &  \thead{85.4} & \textbf{\thead{86.1}} \\
\thead{Name of second accident's \\ driver} & \thead{84.1} & \thead{\textbf{87.1}} & \thead{84.9} & \thead{85.2} \\
\thead{Underwriter} & \thead{92.2} & \thead{92.2} & \thead{92.3} & \thead{\textbf{92.8}} \\
\thead{Responsibility of the first\\ driver of the accident} & \thead{89.8} & \thead{96.5} &  \thead{95.7} & \thead{\textbf{97.2}} \\
\thead{Responsibility of the second\\ driver of the accident} & \thead{89.5} & \thead{95} & \thead{94} & \thead{\textbf{95.3}} \\
\hline
\thead{Mean F1 (\%)} & \thead{90.8} & \thead{95.4} & \thead{93.5} & \thead{\textbf{96.2
}} \\
\thead{DAR (\%)} & \thead{58.1} & \thead{74} & \thead{70.6} & \thead{\textbf{77.5}} \\
\hline
\thead{Time (s)} & \thead{4.1+$t^{*}$} & \thead{6.7} & \thead{18.1} & \thead{\textbf{4.4}} \\
\end{tabular} 
\end{table}Regarding the OCR required by the LayoutLM-v3 approach, we use, for both SROIE and CORD datasets, Microsoft Form Recognizer\footnote{https://learn.microsoft.com/en-us/azure/applied-ai-services/form-recognizer/concept-read?view=form-recog-3.0.0} which includes a document-optimized version of Microsoft Read OCR (MS OCR) as its OCR engine. We note that we tried combining a ResNet-50 \cite{Resnet}-based DBNet++ \cite{DB} for text detection and an SVTR \cite{SVTR} model for text recognition and fine-tuned them on the fields of interest of each dataset. However, the obtained results are worse than those obtained with Microsoft Form Recognizer OCR engine. For the Information Statement Dataset, we don't use MS OCR for confidentiality purposes. Instead, we use an in-house OCR fine-tuned on this dataset to reach the best possible performances. Even though the best OCRs are used for each task, LayoutLM-v3 extraction performances are still lower than those of OCR-free models. This proves the superiority of end-to-end architectures over the OCR-dependent approaches for the information extraction task. We note that for Donut, we use the same input resolution as DocParser. For Dessurt, we use a resolution of $1152 \times 768$, which is the resolution used to pre-train the model.

\section{Primary Experiments and Further Investigation \label{abla}}
\subsection{Primary Experiments}
 In all the experiments, the tested architectures were pre-trained on 0.5 Million synthesized documents and fine-tuned on a deskewed version of the SROIE dataset. We report the inference time on a an Intel Xenion W-2235 CPU, as we aim to provide a model suited for low resources scenarios.
  \setlength{\tabcolsep}{6pt}
\begin{table}
\centering
\setlength\extrarowheight{-20pt}
\caption{\textbf{Comparison of different encoder architectures.} The dataset used is a deskewed version of the SROIE dataset. The field-level F1 score is reported. }\label{tabencoders}
\begin{tabular}{l l l l l}
\hline
\thead{\ Encoder architecture}& \thead{\ F1(\%)} \\
\hline
EasyOCR-based encoder& \thead{54}\\
PP-OCRv2-based encoder&\thead{77}\\
Proposed encoder&\thead{81}
\end{tabular}
\end{table}
\subsubsection{On the Encoder's Architecture} The table~\ref{tabencoders} shows a comparison between an EasyOCR\footnote{https://github.com/JaidedAI/EasyOCR/blob/master/easyocr}-based encoder, a PP-OCRv2 \cite{Paddle}-based encoder and our proposed DocParser encoder. Concerning the EasyOCR and PP-OCRv2 based encoders, each one consists of its corresponding OCR's recognition network followed by few convolutional layers that aim to further reduce the feature map size and increase the receptive field. Our proposed encoder surpasses both encoders by a large margin.
\setlength{\tabcolsep}{6pt}
\begin{table}
\centering
\setlength\extrarowheight{-20pt}
\caption{\textbf{The effect of decreasing the width of the feature map in various stages of DocParser's encoder.} The dataset used is a deskewed version of the SROIE dataset. The field-level F1-score and the extraction time per image on an Intel Xenion W-2235 CPU are reported.}\label{finalfmsize}
\begin{tabular}{l l l l l}
\hline
\thead{\ Encoder stages \\ where the feature \\ map width is reduced} &  \thead{Decoder} &\thead{\ Inference time\\ (seconds)}&F1(\%) \\
\hline
\\
(3,4,5) (proposed) & Transformer & \thead{3.5} & \thead{81}\\
(3,4,5) & LSTM + Additive attention & \thead{3.3} & \thead{58}\\
(1,2,3) & Transformer & \thead{2.1} & \thead{69}\\
(1,2,3) & LSTM + Additive attention & \thead{1.9} & \thead{52}\\
No reduction & Transformer & \thead{6.9} & \thead{81}\\
No reduction & LSTM + Additive attention & \thead{6.6} & \thead{81}
\end{tabular}
\end{table}
\subsubsection{On the Feature Map Width Reduction} While encoding the input image, the majority of the text recognition approaches reduce the dimensions of the feature map mainly vertically \cite{SVTR} \cite{text_recognition}. Intuitively, applying this approach for the information extraction task may seem relevant as it allows different characters to be encoded in different feature vectors. Our empirical results, however, show that this may not always be the case. In fact, we experimented with reducing the encoder's feature map width at different stages. As a decoder, we used both a one layer vanilla Transformer decoder and a Long Short-Term Memory (LSTM) \cite{LSTM} coupled with an attention mechanism that uses an additive attention scoring function \cite{additive}. Table~\ref{finalfmsize} shows that reducing the width of the feature map in the early stages affects drastically the model's accuracy and that reducing the width of the feature map in the later stages achieves the the best speed-accuracy trade-off. Table~\ref{finalfmsize} also shows that while the LSTM-based decoder struggles with a reduced width encoder output, the performance of the vanilla Transformer-based decoder remains the same in both cases. This is probably due to the multi-head attention mechanism that makes the Transformer-based decoder more expressive than an LSTM coupled with an attention mechanism.

\subsubsection{On the Tokenizer Choice} In addition to the RoBERTa tokenizer, we also tested a character-level tokenizer. Table~\ref{tok} shows that the RoBERTa tokenizer allows faster decoding while achieving the same performance as the character-level tokenizer. 
\setlength{\tabcolsep}{6pt}
\begin{table}
\centering
\setlength\extrarowheight{-20pt}
\caption{\textbf{Comparison between different tokenization techniques.} The dataset used is a deskewed version of the SROIE dataset. The field-level F1-score and the decoding time per image on an Intel Xenion W-2235 CPU are reported.}\label{tok}
\begin{tabular}{l l l l l}
\hline
\thead{\ Tokenizer}& \thead{\ Decoding inference time (s) }& \thead{\ F1(\%)} \\
\hline
\\
RoBERTa tokenizer & \thead{0.6} & \thead{81}\\
Character-level tokenizer & \thead{1.2} & \thead{81}\\
\end{tabular}
\end{table}
\subsubsection{On the Number of Decoder Layers} Table~\ref{decoderlayers} shows that increasing the number of decoder layers doesn't improve DocParser's performance. Therefore, using one decoder layer is the best choice as it guarantees less computational cost.

\subsubsection{On the Data Augmentation Strategy} Additionally to the adopted augmentation techniques, we experimented with adding different types of blur and noise to the input images for both the pre-training and the fine-tuning. We concluded that this does not improve DocParser's performance. The lack of performance improvement when using blur may be attributed to the fact that the datasets used for evaluating the model do not typically include blurred images. Additionally, it is challenging to accurately create realistic noise, thus making the technique of adding noise to the input images ineffective.
\setlength{\tabcolsep}{6pt}
\begin{table}
\centering
\setlength\extrarowheight{-20pt}
\caption{\textbf{Effect of the number of decoder layers on the performance and the decoding inference time of DocParser.} The dataset used is a deskewed version of the SROIE dataset. The field-level F1-score and the decoding time per image on an Intel Xenion W-2235 CPU are reported.}\label{decoderlayers}
\begin{tabular}{l l l l l}
\hline
\thead{\ Decoder layers}& \thead{\ Decoding inference time (s) }& \thead{\ F1(\%)} \\
\hline
\\
\thead{1} & \thead{0.6} & \thead{81}\\
\thead{2} & \thead{0.8} & \thead{81}\\
\thead{4} & \thead{1.2} & \thead{81}
\end{tabular}
\end{table}

\subsection{Further Investigation}
\setlength{\tabcolsep}{6pt}
\begin{table}
\centering
\setlength\extrarowheight{-20pt}
\caption{\textbf{Comparison between different pre-training strategies.} All the models are pre-trained for a total of 70k steps. The field-level F1-score is reported.}\label{abc}
\begin{tabular}{l l l l l}
\hline
\thead{\ Pre-training tasks}& \thead{\ SROIE } & \thead{\ CORD } & \thead{\ ISD } \\
\hline
\\
Knowledge transfer  & \thead{77.5} & \thead{75} & \thead{89.7} \\
Knowledge transfer + Document reading & \thead{84.7} & \thead{83.7} & \thead{95.6} \\
Knowledge transfer + Masked document reading & \textbf{\thead{84.9}} & \textbf{\thead{84.2}} & \textbf{\thead{95.9}}
\end{tabular}
\end{table}
\subsubsection{On the Pre-training Strategy} Table \ref{abc} presents a comparison between different pre-training strategies. To reduce compute used, all the models were pre-trained for 70k back-propagation steps, with 7k knowledge transfer steps in the case of two pre-training tasks. The results show that masking text regions during the document reading pre-training task does effectively lead to an increase in performance on all three datasets. It also confirms, as demonstrated in \cite{donut} and \cite{Dessurt}, that document reading, despite its simplicity, is an effective pre-training task.

\subsubsection{On the Input Resolution} 
\begin{figure}[h]
\caption{\textbf{The impact of the input resolution on DocParser's performance on the SROIE dataset.} The field-level F1 score is reported.}\label{resolutionimpact}
\centering
\includegraphics[width=0.75\textwidth]{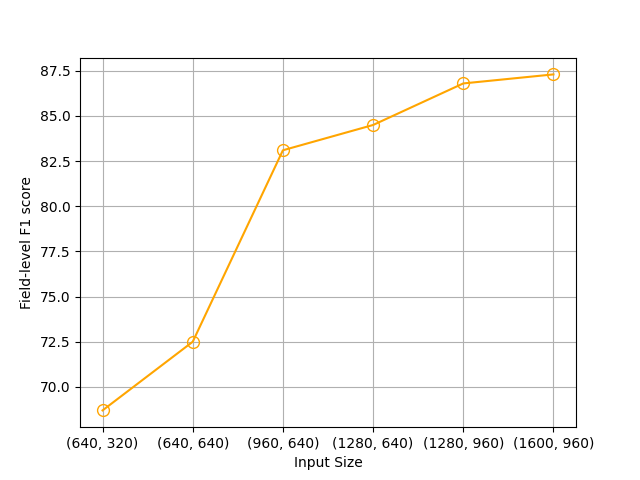}
\end{figure}

Figure~\ref{resolutionimpact} shows the effect of the input resolution on the performance of DocParser on the SROIE dataset. DocParser shows satisfying results even with a low-resolution input. It achieves 83.1 field-level F1 score with a $960 \times 640$ input resolution. The inference time for this resolution  on an Intel Xenion W-2235 CPU is only 1.7 seconds. So, even at this resolution, DocParser still surpasses Donut and LayoutLM-v3 on SROIE while being more than three times faster.
However, if the input resolution is set to $640 \times 640$ or below, the model's performance shows a drastic drop. This may be due to the fact that the characters start to be illegible at such a low resolution.

\section{Conclusion}

We have introduced DocParser, a fast end-to-end approach for information extraction from visually rich documents. Contrary to previously proposed end-to-end models, DocParser's encoder is specifically designed to capture both intra-character local patterns and
inter-character long-range dependencies. Experiments on both public and private datasets showed that DocParser achieves state-of-the-art results in terms of both speed and accuracy which makes it perfectly suitable for real-world applications.

\subsubsection{Acknowledgments} 
The authors wish to convey their genuine appreciation to Prof. Davide Buscaldi and Prof. Sonia Vanier for providing them with valuable guidance. Furthermore, the authors would like to express their gratitude to Paul Wassermann  and Arnaud Paran  for their assistance in proofreading previous versions of the manuscript.

%
%
{
\small
\bibliographystyle{splncs04}
\bibliography{egbib}

\begin{thebibliography}{10}
\providecommand{\url}[1]{\texttt{#1}}
\providecommand{\urlprefix}{URL }
\providecommand{\doi}[1]{https://doi.org/#1}

\bibitem{text_recognition}
Baek, J., Kim, G., Lee, J., Park, S., Han, D., Yun, S., Oh, S.J., Lee, H.: What
  is wrong with scene text recognition model comparisons? dataset and model
  analysis. In: Proceedings of the IEEE/CVF international conference on
  computer vision. pp. 4715--4723 (2019)

\bibitem{text_detection}
Baek, Y., Lee, B., Han, D., Yun, S., Lee, H.: Character region awareness for
  text detection. In: Proceedings of the IEEE/CVF conference on computer vision
  and pattern recognition. pp. 9365--9374 (2019)

\bibitem{earlier_approaches_2}
Cesarini, F., Francesconi, E., Gori, M., Soda, G.: Analysis and understanding
  of multi-class invoices. Document Analysis and Recognition  \textbf{6},
  102--114 (2003)

\bibitem{trie++}
Cheng, Z., Zhang, P., Li, C., Liang, Q., Xu, Y., Li, P., Pu, S., Niu, Y., Wu,
  F.: Trie++: Towards end-to-end information extraction from visually rich
  documents. arXiv preprint arXiv:2207.06744  (2022)

\bibitem{Dessurt}
Davis, B., Morse, B., Price, B., Tensmeyer, C., Wigington, C., Morariu, V.:
  End-to-end document recognition and understanding with dessurt. In: Computer
  Vision--ECCV 2022 Workshops: Tel Aviv, Israel, October 23--27, 2022,
  Proceedings, Part IV. pp. 280--296. Springer (2023)

\bibitem{Bert_grid}
Denk, T.I., Reisswig, C.: Bertgrid: Contextualized embedding for 2d document
  representation and understanding. In Workshop on Document Intelligence at
  NeurIPS.  (2019)

\bibitem{Bert}
Devlin, J., Chang, M.W., Lee, K., Toutanova, K.: {BERT}: Pre-training of deep
  bidirectional transformers for language understanding. In: Proceedings of the
  2019 Conference of the North {A}merican Chapter of the Association for
  Computational Linguistics: Human Language Technologies, Volume 1 (Long and
  Short Papers). pp. 4171--4186. Association for Computational Linguistics,
  Minneapolis, Minnesota (Jun 2019)

\bibitem{SVTR}
Du, Y., Chen, Z., Jia, C., Yin, X., Zheng, T., Li, C., Du, Y., Jiang, Y.G.:
  Svtr: Scene text recognition with a single visual model. In: Raedt, L.D.
  (ed.) Proceedings of the Thirty-FirstInternational Joint Conference on
  Artificial Intelligence, {IJCAI-22}. pp. 884--890. International Joint
  Conferences on Artificial Intelligence Organization (7 2022), main Track

\bibitem{Paddle}
Du, Y., Li, C., Guo, R., Cui, C., Liu, W., Zhou, J., Lu, B., Yang, Y., Liu, Q.,
  Hu, X., Yu, D., Ma, Y.: Pp-ocrv2: Bag of tricks for ultra lightweight ocr
  system. ArXiv  \textbf{abs/2109.03144} (2021)

\bibitem{Lambert}
Garncarek, {\L}., Powalski, R., Stanis{\l}awek, T., Topolski, B., Halama, P.,
  Turski, M., Grali{\'{n} }ski, F.: {LAMBERT}: Layout-aware language modeling
  for information extraction. In: Document Analysis and Recognition {ICDAR}
  2021, pp. 532--547. Springer International Publishing (2021)

\bibitem{LSTM}
Graves, A., Graves, A.: Long short-term memory. Supervised sequence labelling
  with recurrent neural networks pp. 37--45 (2012)

\bibitem{eaten}
Guo, H., Qin, X., Liu, J., Han, J., Liu, J., Ding, E.: Eaten: Entity-aware
  attention for single shot visual text extraction. In: 2019 International
  Conference on Document Analysis and Recognition (ICDAR). pp. 254--259 (2019)

\bibitem{Resnet}
He, K., Zhang, X., Ren, S., Sun, J.: Deep residual learning for image
  recognition. In: Proceedings of the IEEE conference on computer vision and
  pattern recognition. pp. 770--778 (2016)

\bibitem{Bros}
Hong, T., Kim, D., Ji, M., Hwang, W., Nam, D., Park, S.: Bros: A pre-trained
  language model focusing on text and layout for better key information
  extraction from documents. In: Proceedings of the AAAI Conference on
  Artificial Intelligence. vol.~36, pp. 10767--10775 (2022)

\bibitem{mobilenet}
Howard, A.G., Zhu, M., Chen, B., Kalenichenko, D., Wang, W., Weyand, T.,
  Andreetto, M., Adam, H.: Mobilenets: Efficient convolutional neural networks
  for mobile vision applications (2017)

\bibitem{stocha}
Huang, G., Sun, Y., Liu, Z., Sedra, D., Weinberger, K.Q.: Deep networks with
  stochastic depth. In: Computer Vision--ECCV 2016: 14th European Conference,
  Amsterdam, The Netherlands, October 11--14, 2016, Proceedings, Part IV 14.
  pp. 646--661. Springer (2016)

\bibitem{layoutlm}
Huang, Y., Lv, T., Cui, L., Lu, Y., Wei, F.: Layoutlmv3: Pre-training for
  document ai with unified text and image masking. In: Proceedings of the 30th
  ACM International Conference on Multimedia. p. 4083–4091. MM '22,
  Association for Computing Machinery, New York, NY, USA (2022)

\bibitem{chargrid}
Katti, A.R., Reisswig, C., Guder, C., Brarda, S., Bickel, S., H{\"o}hne, J.,
  Faddoul, J.B.: {C}hargrid: Towards understanding 2{D} documents. In:
  Proceedings of the 2018 Conference on Empirical Methods in Natural Language
  Processing. pp. 4459--4469. Association for Computational Linguistics,
  Brussels, Belgium (Oct-Nov 2018)

\bibitem{visualwordgrid}
Kerroumi, M., Sayem, O., Shabou, A.: Visualwordgrid: Information extraction
  from scanned documents using a multimodal approach. In: Document Analysis and
  Recognition--ICDAR 2021 Workshops: Lausanne, Switzerland, September 5--10,
  2021, Proceedings, Part II. pp. 389--402. Springer (2021)

\bibitem{donut}
Kim, G., Hong, T., Yim, M., Nam, J., Park, J., Yim, J., Hwang, W., Yun, S.,
  Han, D., Park, S.: Ocr-free document understanding transformer. In: Avidan,
  S., Brostow, G., Ciss{\'e}, M., Farinella, G.M., Hassner, T. (eds.) Computer
  Vision -- ECCV 2022. pp. 498--517. Springer Nature Switzerland, Cham (2022)

\bibitem{OCR_Post_Correction_2}
Kissos, I., Dershowitz, N.: Ocr error correction using character correction and
  feature-based word classification. In: 2016 12th IAPR Workshop on Document
  Analysis Systems (DAS). pp. 198--203 (2016)

\bibitem{docreader}
Klaiman, S., Lehne, M.: Docreader: bounding-box free training of a document
  information extraction model. In: Document Analysis and Recognition--ICDAR
  2021: 16th International Conference, Lausanne, Switzerland, September 5--10,
  2021, Proceedings, Part I 16. pp. 451--465. Springer (2021)

\bibitem{VIT}
Kolesnikov, A., Dosovitskiy, A., Weissenborn, D., Heigold, G., Uszkoreit, J.,
  Beyer, L., Minderer, M., Dehghani, M., Houlsby, N., Gelly, S., Unterthiner,
  T., Zhai, X.: An image is worth 16x16 words: Transformers for image
  recognition at scale. In: International Conference on Learning
  Representations (2021)

\bibitem{CDIP}
Lewis, D., Agam, G., Argamon, S., Frieder, O., Grossman, D., Heard, J.:
  Building a test collection for complex document information processing. In:
  Proceedings of the 29th Annual International ACM SIGIR Conference on Research
  and Development in Information Retrieval. p. 665–666. SIGIR '06,
  Association for Computing Machinery, New York, NY, USA (2006)

\bibitem{Bart}
Lewis, M., Liu, Y., Goyal, N., Ghazvininejad, M., Mohamed, A., Levy, O.,
  Stoyanov, V., Zettlemoyer, L.: {BART}: Denoising sequence-to-sequence
  pre-training for natural language generation, translation, and comprehension.
  In: Proceedings of the 58th Annual Meeting of the Association for
  Computational Linguistics. pp. 7871--7880. Association for Computational
  Linguistics, Online (Jul 2020)

\bibitem{DB}
Liao, M., Zou, Z., Wan, Z., Yao, C., Bai, X.: Real-time scene text detection
  with differentiable binarization and adaptive scale fusion. IEEE Transactions
  on Pattern Analysis and Machine Intelligence  \textbf{45}(1),  919--931
  (2022)

\bibitem{FPN}
Lin, T.Y., Doll{\'a}r, P., Girshick, R., He, K., Hariharan, B., Belongie, S.:
  Feature pyramid networks for object detection. In: Proceedings of the IEEE
  conference on computer vision and pattern recognition. pp. 2117--2125 (2017)

\bibitem{Graph_based-1}
Liu, X., Gao, F., Zhang, Q., Zhao, H.: Graph convolution for multimodal
  information extraction from visually rich documents. In: Proceedings of the
  2019 Conference of the North {A}merican Chapter of the Association for
  Computational Linguistics: Human Language Technologies, Volume 2 (Industry
  Papers). pp. 32--39. Association for Computational Linguistics, Minneapolis,
  Minnesota (Jun 2019)

\bibitem{RoBERTa}
Liu, Y., Ott, M., Goyal, N., Du, J., Joshi, M., Chen, D., Levy, O., Lewis, M.,
  Zettlemoyer, L., Stoyanov, V.: Roberta: A robustly optimized bert pretraining
  approach. arXiv preprint arXiv:1907.11692  (2019)

\bibitem{Swin}
Liu, Z., Lin, Y., Cao, Y., Hu, H., Wei, Y., Zhang, Z., Lin, S., Guo, B.: Swin
  transformer: Hierarchical vision transformer using shifted windows. In:
  Proceedings of the IEEE/CVF international conference on computer vision. pp.
  10012--10022 (2021)

\bibitem{ConvNext}
Liu, Z., Mao, H., Wu, C.Y., Feichtenhofer, C., Darrell, T., Xie, S.: A convnet
  for the 2020s. In: Proceedings of the IEEE/CVF Conference on Computer Vision
  and Pattern Recognition. pp. 11976--11986 (2022)

\bibitem{cosine}
Loshchilov, I., Hutter, F.: Sgdr: Stochastic gradient descent with warm
  restarts. in iclr, 2017. arXiv preprint arXiv:1608.03983  (2016)

\bibitem{ADAMW}
Loshchilov, I., Hutter, F.: Decoupled weight decay regularization. In:
  International Conference on Learning Representations (2017)

\bibitem{earlier_approaches_1}
Medvet, E., Bartoli, A., Davanzo, G.: A probabilistic approach to printed
  document understanding. Int. J. Doc. Anal. Recognit.  \textbf{14}(4),
  335–347 (dec 2011)

\bibitem{cloudscan}
Palm, R.B., Winther, O., Laws, F.: Cloudscan-a configuration-free invoice
  analysis system using recurrent neural networks. In: 2017 14th IAPR
  International Conference on Document Analysis and Recognition (ICDAR).
  vol.~1, pp. 406--413. IEEE (2017)

\bibitem{TILIT}
Powalski, R., Borchmann, {\L}., Jurkiewicz, D., Dwojak, T., Pietruszka, M.,
  Pa{\l}ka, G.: Going full-tilt boogie on document understanding with
  text-image-layout transformer. In: Document Analysis and Recognition--ICDAR
  2021: 16th International Conference, Lausanne, Switzerland, September 5--10,
  2021, Proceedings, Part II 16. pp. 732--747. Springer (2021)

\bibitem{earlier_approaches_0}
Rusiñol, M., Benkhelfallah, T., dAndecy, V.P.: Field extraction from
  administrative documents by incremental structural templates. In: 2013 12th
  International Conference on Document Analysis and Recognition. pp. 1100--1104
  (2013)

\bibitem{OCR_Post_Correction}
Schaefer, R., Neudecker, C.: A two-step approach for automatic {OCR}
  post-correction. In: Proceedings of the The 4th Joint SIGHUM Workshop on
  Computational Linguistics for Cultural Heritage, Social Sciences, Humanities
  and Literature. pp. 52--57. International Committee on Computational
  Linguistics, Online (Dec 2020)

\bibitem{Graph_based-2}
Sun, H., Kuang, Z., Yue, X., Lin, C., Zhang, W.: Spatial dual-modality graph
  reasoning for key information extraction. arXiv preprint arXiv:2103.14470
  (2021)

\bibitem{attention}
Vaswani, A., Shazeer, N., Parmar, N., Uszkoreit, J., Jones, L., Gomez, A.N.,
  Kaiser, {\L}., Polosukhin, I.: Attention is all you need. Advances in neural
  information processing systems  \textbf{30} (2017)

\bibitem{VIES}
Wang, J., Liu, C., Jin, L., Tang, G., Zhang, J., Zhang, S., Wang, Q., Wu, Y.,
  Cai, M.: Towards robust visual information extraction in real world: New
  dataset and novel solution. In: Proceedings of the AAAI Conference on
  Artificial Intelligence. vol.~35, pp. 2738--2745 (2021)

\bibitem{additive}
Wang, W., Yang, N., Wei, F., Chang, B., Zhou, M.: Gated self-matching networks
  for reading comprehension and question answering. In: Proceedings of the 55th
  Annual Meeting of the Association for Computational Linguistics (Volume 1:
  Long Papers). pp. 189--198. Association for Computational Linguistics,
  Vancouver, Canada (Jul 2017)

\bibitem{Layout-aware}
Wei, M., He, Y., Zhang, Q.: Robust layout-aware ie for visually rich documents
  with pre-trained language models. In: Proceedings of the 43rd International
  ACM SIGIR Conference on Research and Development in Information Retrieval.
  pp. 2367--2376 (2020)

\bibitem{Resnext}
Xie, S., Girshick, R., Doll{\'a}r, P., Tu, Z., He, K.: Aggregated residual
  transformations for deep neural networks. In: Proceedings of the IEEE
  conference on computer vision and pattern recognition. pp. 1492--1500 (2017)

\bibitem{layoutlmv1}
Xu, Y., Xu, Y., Lv, T., Cui, L., Wei, F., Wang, G., Lu, Y., Florencio, D.,
  Zhang, C., Che, W., Zhang, M., Zhou, L.: {L}ayout{LM}v2: Multi-modal
  pre-training for visually-rich document understanding. In: Proceedings of the
  59th Annual Meeting of the Association for Computational Linguistics and the
  11th International Joint Conference on Natural Language Processing (Volume 1:
  Long Papers). pp. 2579--2591. Association for Computational Linguistics,
  Online (Aug 2021)

\bibitem{layoutlmv0}
Xu, Y., Li, M., Cui, L., Huang, S., Wei, F., Zhou, M.: {LayoutLM}: Pre-training
  of text and layout for document image understanding. In: Proceedings of the
  26th {ACM} {SIGKDD} International Conference on Knowledge Discovery and Data
  Mining. {ACM} (aug 2020)

\bibitem{Cutie}
Zhao, X., Niu, E., Wu, Z., Wang, X.: Cutie: Learning to understand documents
  with convolutional universal text information extractor. arXiv preprint
  arXiv:1903.12363  (2019)

\end{thebibliography}
}
\end{document}